# Design and Control of a Quasi-Direct Drive Soft Exoskeleton for Knee Injury Prevention during Squatting

Shuangyue Yu[*], Tzu-Hao Huang[*], Dianpeng Wang, Brian Lynn, Dina Sayd, Viktor Silivanov, Young Soo Park, Yingli Tian, Fellow, *IEEE*, and Hao Su[†], Member, *IEEE*

*Abstract*—This paper presents design and control innovations of wearable robots that tackle two barriers to widespread adoption of powered exoskeletons, namely restriction of human movement and versatile control of wearable co-robot systems. First, the proposed quasi-direct drive actuation comprising of our customized high-torque density motors and low ratio transmission mechanism significantly reduces the mass of the robot and produces high backdrivability. Second, we derive a biomechanics model-based control that generates biological torque profile for versatile control of both squat and stoop lifting assistance. **The control algorithm detects lifting postures using compact inertial measurement unit (IMU) sensors to generate an assistive profile that is proportional to the biological torque produced from our model. Experimental results demonstrate that the robot exhibits low mechanical impedance (1.5 Nm resistive torque) when it is unpowered and 0.5 Nm resistive torque with zero-torque tracking control. Root mean square (RMS) error of torque tracking is less than 0.29 Nm (1.21% error of 24 Nm peak torque).** Compared with squatting without the exoskeleton, the controller reduces 87.5%, 80% and 75% of the of three knee extensor muscles (average peak EMG of 3 healthy subjects) during squat with 50% of biological torque assistance.

## I. INTRODUCTION

Musculoskeletal disorders (MSDs) are a leading cause of injury among various individuals, [1]. It is estimated that the direct costs of injuries due to overexertion from lifting, pushing, pulling, turning, throwing, or catching to be $15.1 billion in 2016 [1]. Wearable robots present an attractive solution to mitigate the incidence of injury and augment human performance [2]. Besides recent breakthroughs of wearable robotics in human augmentation that enhance the walking economy and endurance [3-5] and in gait restoration that enhances mobility [6-9], industrial exoskeletons are an emerging area that presents new opportunities and challenges. Passive [10, 11] and powered exoskeletons [12, 13] have demonstrated effectiveness for injury prevention of upper body and back support. Recently, there is a growing interest in wearable robots for knee joint assistance as cumulative knee disorders account for 65% of lower extremity musculoskeletal disorders [14]. Squatting and kneeling are two of the primary risk factors that contribute to knee disorders [14].

Like all wearable robots, knee exoskeletons can be generally classified as rigid or soft in terms of actuation and transmission. First, it is recognized that excessive mass and high impedance are two key drawbacks of state of the art wearable robots [15]. Quasi-passive knee design was studied in [16] as prosthesis and [17] as exoskeletons [18] using series elastic actuators (SEA) to decouple motor inertia and reduce passive output impedance. Keeogo exoskeleton [15] uses high ratio harmonic gear to amplify torque of a brushless direct current (BLDC) motor. Second, most of the existing knee exoskeletons are designed for walking assistance [19, 20] and they typically do not allow squat motion due to the interference between the robot structure and human bodies (e.g. [21, 22]).

Soft exoskeletons using either pneumatics [23] or cable transmission [24] represent a trend in wearable robot design. Pneumatic actuation operates on tethered air compressor [23], thus it is challenging for portable system applications. Textile soft exosuit is a new approach of soft wearable robot design and has been used for ankle [25] and hip joint [3] assistance during walking. There is no knee textile exosuit developed yet, possibly due to the demand to anchor wearable structures to thigh and shank while the ankle and hip exosuits can be anchored to footwear and waist respectively. Squatting motion is relatively simpler than walking as it involves fewer muscle groups, but its functional requirements present new challenges because it needs to overcome the same limitations, while the range of motion and the torque assistance during squatting are much greater than walking, as shown in Table.1.

To overcome the limitations of restriction of natural movement and versatile control of human-robot interaction [2, 26], the contributions of this work include: 1) a quasi-direct drive actuation paradigm using our high torque density motors

This work is supported by the National Science Foundation grant NRI 1830613 and Grove School of Engineering, The City University of New York, City College. Any opinions, findings, and conclusions or recommendations expressed in this material are those of the author (s) and do not necessarily reflect the views of the funding organizations.

S. Yu, T. Huang, D. Wang, B. Lynn, D. Sayd, V. Silivanov, and H. Su are with the Lab of Biomechatronics and Intelligent Robotics, Department of Mechanical Engineering, The City University of New York, City College, New York, NY, 10023, US. Y.S. Park is with the Applied Materials Division, Argonne National Laboratory, Lemont, IL, 60439, US. Y. Tian is with the Department of Electrical Engineering, The City University of New York, City College, New York, NY, 10023, US. *These authors contributed equally to this work. †Corresponding author. Email: hao.su@ccny.cuny.edu

that significantly reduce the mass and mechanical impedance of the wearable robot; and 2) a biomechanics model based versatile control strategy that generates a unified biological torque profile to assist both squat and stoop lifting activities. This paper demonstrates the potential of our solution with brushless direct current (DC) motors for portable systems in comparison with the alternating current (AC) motor based tethered exoskeletons [22] and improvement of backdrivability in comparison with [3] thanks to higher torque density of our motors than other motors (Allied Motion Technologies) used in [3] (Section II). A knee exoskeleton, as shown in Fig. 1, is instantiated as one example of our actuation paradigm, but the presented innovation is generic for design and control of a wide variety of high-performance wearable robots.

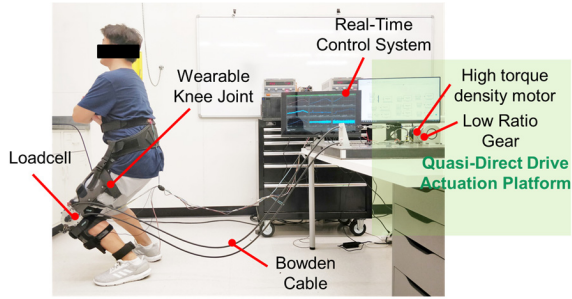

Fig. 1. A healthy subject performs squatting with the soft exoskeleton that uses soft cable transmission with rigid customized wearable structure. The quasi-direct drive actuation comprising of high torque density motors and low ratio transmission makes the exoskeleton highly-backdrivable.

## II. DESIGN REQUIREMENTS

The design requirement as shown in Table 1 is based on data from healthy human subjects (80 kg weight and 180 cm height) without carrying any loads [27]. Knee joint assistance during squatting necessitates a broad range of motion (0-130° flexion) and biological joint moment (up to 60 Nm). The torque generated from the robot needs to be delivered at an angular velocity of no less than 2.4 rad/s to effectively synchronize with wearers. The robot is designed to deliver 72 Nm torque to provide more than 50% of biological lower limb joints as our control philosophy is to use small to medium levels of torque in combination with optimized timing, magnitude, and profile of torque trajectories [3].

TABLE I. DESIGN PARAMETERS OF KNEE EXO FOR DEEP SQUAT

| Parameters | Walking | Squat | Our Robot |
|---|---|---|---|
| Range of motion (deg) | 10-60 | 0-130 | 0-130 |
| Max knee joint moment (Nm) | 40 | 60 | 72 |
| Max knee joint speed (rad/s) | 4.3 | 2.4 | 4.4 |
| Exoskeleton weight (kg) | —— | —— | 1.1 |
| Actuator min speed (m/s) | 0.22 | 0.12 | 0.22 |
| Actuator max force (N) | 320 | 480 | 1250 |

## III. QUASI-DIRECT DRIVE ACTUATION USING HIGH TORQUE DENSITY MOTORS

Quasi-direct drive actuation [28] [29] is a new paradigm of robot actuation design that leverages high torque density motors with low ratio transmission mechanism. In this paper, the quasi-direct drive is a generalized definition to describe actuation with low ratio transmission mechanism because of high torque density motors without necessarily referring to motors directly attached to mechanical loads as its conventional definition. It has been recently studied for legged robots [28] and exoskeletons [30]. To our knowledge, this paper is the first work to investigate quasi-direct drive and its feasibility for wearable knee co-robots to augment squatting movement. The benefits of the quasi-direct drive include a simplified mechanical structure, reduced mass and volume, and highly backdrivability. Thus, it is ideal for wearable robots in terms of static and dynamic requirements. Our soft knee exoskeleton is a versatile assistive device to augment knee movements during lifting (both squat and stoop) and walking, though the focus of this paper is squat assistance. Since the focus of this work is to understand the feasibility of the design principles and effectiveness of control strategies, the prototype is a tethered system with offboard actuation.

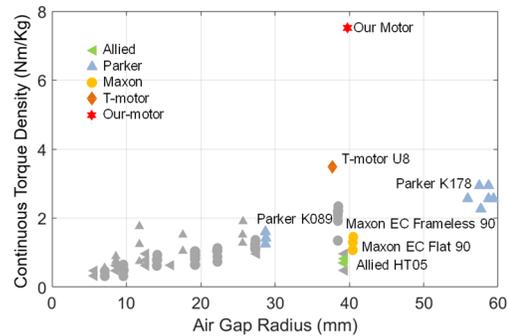

Fig. 2. Continuous torque density versus air gap radius distribution of our motor and other commercial ones. Exoskeleton design typically needs motors with air gap radius in the 35-40 mm range. Our custom designed motor marked as a star has 5 times continuous torque density than Maxon brushless DC motor EC flat 90 that is widely used in exoskeleton industry.

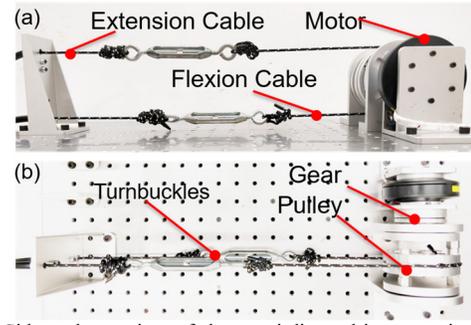

Fig. 3. Side and top view of the quasi-direct drive actuation platform consisting of high torque density motors and low transmission ratio gears (36:1 custom designed planetary gear). The turnbuckles are a tensioner mechanism to apply pre-loading force to the cable transmission. Compared with [3] that use one motor to actuate a unidirectional motion, our mechanism uses one motor to generate bidirectional actuation, thus it can further reduce the mass of the overall system.

To enable the quasi-direct drive actuation paradigm, it is crucial to design high torque density motors. Our custom designed BLDC motors optimize the mechanical structure, topology, and electromagnetic properties [31]. It uses high-temperature resistive magnetic materials and adopts an outer rotor, flat and concentrated winding structure. Unlike conventional BLDC motors that place windings around rotors, our rotor consists of only the permanent magnet and rotor cover while the winding is attached to stators. This design significantly reduces the inertia and mechanical impedance of the motor while increasing its control bandwidth. Fig. 2 shows the continuous torque density versus air gap radius distribution of our motor and commercial ones [29]. The continuous torque of our motor is 2 Nm and its mass is 274 g. Maxon EC Flat 90 has 0.5 Nm continuous torque with 648 g mass. In the 35-40 mm air gap radius domain, the continuous torque density of our motor is 7.3 Nm/Kg while T-motor U8 is 3.5 Nm/Kg, and Maxon EC Flat 90 is 1.5 Nm/Kg. The quasi-direct drive actuation paradigm was implemented with a tethered actuation platform shown in Fig. 3. Using 2-stage planetary gears with 36:1 ratio and 290 g mass, the actuator generates 72 Nm continuous torque and 4.36 rad/s angular velocity.

## IV. ELECTROMECHANICAL DESIGN OF A SOFT EXOSKELETON

The soft exoskeleton design approach proposed in this paper uses a cable transmission (like textile-based soft exosuit) in combination with a rigid wearable structure with interior soft padding (like rigid exoskeletons producing large torque). Our hybrid soft exoskeleton has a larger moment arm (distance between human joint and the lumped center of the wearable structure, the same as rigid exoskeletons [15] [18]) than textile soft exosuits [3] (approximately the radius of the attached limb) and avoids shear forces to human [25]. Thus, the hybrid soft exoskeleton requires much less force from the cable system to deliver the same amount of torque than textile soft exosuits. It presents one solution to reduce forces applied to limbs (because of its large moment arm) and pressure concentration (3D scanning and 3D printing based orthotic brace with foam padding are conformable and conformal vs. textile interface [25]). The soft exoskeleton is implemented with the quasi-direct drive actuation, a bidirectional Bowden cable transmission mechanism, and a low-profile knee joint mechanism. Though the current platform is configured as a tethered system, it can be converted to a portable system, as the overall mass of motor and gears are 274g and 290 g respectively. Moreover, our mechanism design further reduces system mass by a bidirectional Bowden cable transmission mechanism (similar to [32] and [33]) that uses one motor to generate bidirectional actuation instead of one motor controls unidirectional motion in textile soft exosuits.

### A. Low Profile and Lightweight Knee Joint Mechanism

The design consideration of knee joint mechanism is to avoid interference with the human body during squat motion while achieving minimal mass. The knee joint shown in Fig. 4 is the distal portion of the bidirectional cable-drive mechanism. The design includes one flexion cable and one extension cable that pass through the distal pulley and terminate at the cable locking mechanism. The load cell connects the thigh and calf braces and plays a key role in force transmission between the cable and the shank plates. The top portion of the knee mechanism is attached to the 3D printed thigh brace while the shank plate is fixed to the calf brace. The terminated cable on the locking mechanism actuates the pulley, thus driving the shank plate via a load cell.

### B. Customized Low-Cost Wearable Structure

The exoskeleton is attached to the body via 3D carbon fiber printed braces designed to conform to the human leg. These braces transmit the torque at the pulley knee joint into a pressure distributed along the length of the thigh and shank.

Therefore, the appropriate size of wearable arms plays a crucial role in the performance and comfort of the subject wearing the exoskeleton. Three-dimensional infrared scans (Sense 2, MatterHackers Inc.) are taken of the patient's leg and then processed into a three-dimensional CAD model, which are 3D printed using fused deposition modeling. This model is then reinforced with a carbon fiber and resin composite. The arms are padded in locations of leg contact to aid in comfort. Velcro straps are then wrapped around the leg of the user and are anchored to the exoskeleton arms, thus allowing the exoskeleton to be adjusted for optimal user comfort.

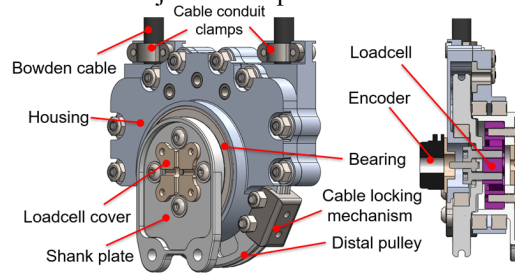

Fig. 4. A section view (left) and an isometric view (right) of the cable-driven knee mechanism that actuates both knee flexion and extension. It is composed of two cable conduit clamps, a distal pulley, a housing to enclose internal components, a load cell cover, a shank plate, a cable locking mechanism (it also prevents knee hyperextension), a ball bearing, an encoder, and a custom load cell that measures up to 50 Nm torque.

### C. Electronics and Communication

The electronics system has a two-level configuration architecture: real-time target computer as a high-level controller and local motor driver electronics as a low-level controller. The high-level controller uses a desktop computer to run MATLAB Simulink Real-Time and executes the real-time control algorithm. The low-level controller can measure motor status (i.e. current, velocity, and position) in real-time and communicate with the target computer through CAN bus. Besides, three IMU sensors, five EMG sensors, and one loadcell were connected to the desktop computer through corresponding interface boards.

## V. SQUAT ASSISTIVE CONTROL STRATEGIES

We derive a biomechanics model that explicitly generates a unified biological torque profile to assist both squat and stoop lifting activities in real time. Unlike methods that use simple and pre-defined profiles (e.g. sine waves) to approximate biological joint torque, this novel method is biologically meaningful and applicable to squat, stoop and walking activities. [34] proposed assistive algorithms for a squat assistance exoskeleton. But the model assumed that the back of the subject was straight, and the trunk angle was zero. It only used knee joint to calculate the torque reference and lacks the posture information of the hip and trunk. During lifting (squat and stoop) the back angle varies and it significantly affects the knee joint torque.

The assistive control as shown in Fig. 5 is composed of high-level torque control and low-level motor control following the method in Roy et al. [35] that demonstrated force tracking of the robot arm in contact with surfaces of unknown linear compliance by the force control with inner loop velocity control. It proved that its controller guarantees arbitrarily small force errors for bounded inner loop velocity tracking errors.

Since this paper does not focus on impedance regulation, we adapted the method by Roy et al. [35] due to its simplicity and feasibility to control the interaction torque between the exoskeleton and human (similar to control the interaction force between the robot arm and environment in the applications of [35]) as demonstrated in Section VI.

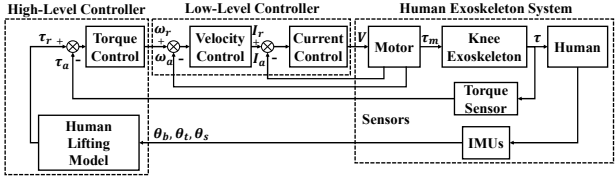

Fig. 5. The diagram of the bio-inspired lifting assistance control algorithm. The high-level controller generates a reference torque profile using our generic lifting biomechanics model. The $\tau_r, \tau_a, \tau,$ and $\tau_m$ are the torque reference, actual measured torque, the actual assistive torque, and the output torque of motor; the $\omega_r$ and $\omega_a$ are the velocity reference and actual estimated velocity; the $I_r$ and $I_a$ are the current reference and actual measured current. The $V$ is the applied voltage for the motor. The $\theta_b$, $\theta_t$, and $\theta_s$ are the trunk angle, thigh angle, and shank angle.

*A. Human Quasi-Static Model during Squat*

A human biomechanics model as shown in Fig. 6 is derived to calculate the biological knee joint torque and assistive torque. This model is customizable to different individuals since the calculated torque can be adjusted according to the subject's weight and height through the weight ratio (subject weight $M_{sb}$/human model weight $M_W$) and height ratio (subject height $L_{sb}$/human model height $L_H$). Since squat and stoop involves significantly different biomechanics of the knee joint, our model is versatile in the sense that it can cover both scenarios for a wide variety of people. The knee joint torque ($\tau_k$) can be derived from equation (1). It works with both fast and slow motions in lifting tasks.

$$\tau_k = I(\theta)\ddot{\theta} + C(\theta, \dot{\theta}) + G(\theta) \qquad (1)$$

where $\theta$ is the joint angles, $I(\theta)$ is the inertia matrix, $C(\theta, \dot{\theta})$ is the centrifugal and Coriolis loading, and $G(\theta)$ is the gravitational loading.

Because lifting tasks are typically relatively slow, and the knee joint torque is dominated by the gravitational loading. Thus estimated knee joint torque ($\hat{\tau}_k$) by computed by a quasi-static model expressed in equation (2).

$$\hat{\tau}_k = G(\theta) = -0.5 \cdot [M_b \cdot g \cdot (L_b \cdot \sin\theta_b + L_t \cdot \sin\theta_t) + M_t \cdot g \cdot L_{tc} \cdot \sin\theta_t] \qquad (2)$$

Here the knee extension is defined as the positive direction for the knee joint torque $\tau_k$ and reference torque $\tau_r$. The clockwise direction is defined as the positive direction for the trunk angle $\theta_b$, the thigh angle $\theta_t$, and the shank angle $\theta_s$. $M_b$ is the combined mass of the head, neck, thorax, abdomen, pelvis, arms, forearms, and hands, and $M_t$ is the mass of thigh, $L_b$ is the length between the center of mass of $M_b$ and the hip pivot, whereas $L_t$ is the length of thigh between the hip pivot and knee pivot, $L_{tc}$ is the length between the center of mass of $M_t$ and the knee pivot, $g$ is the gravitational constant, $\theta_b$ is the trunk angle and $\theta_t$ is the thigh angle. The desired assistive torque of the exoskeleton ($\tau_r$) was defined as equation (3) in our proposed assistive control.

$$\tau_r = \alpha \cdot \hat{\tau}_k \qquad (3)$$

As long as the gain $\alpha$ is positive, the exoskeleton will assist the human. It can be used to reduce the loading and increase the endurance of workers. On the other hand, when the gain $\alpha$ is negative, the exoskeleton will resist the human. It can be used to increase the muscle strength for healthy subjects in fitness or individuals with movement impairments in rehabilitation.

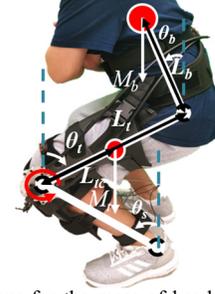

Fig. 6. The annotations for the mass of head, neck, thorax, abdomen, pelvis, arms, forearms, and hands ($M_b$), the mass of thigh ($M_t$), the length between the center of mass of $M_t$ and the hip pivot ($L_b$), the length of thigh between the hip pivot and knee pivot ($L_t$), the length between the center of mass of $M_t$ and the knee pivot ($L_{tc}$), the trunk angle ($\theta_b$), the thigh angle ($\theta_t$), and the shank angle ($\theta_s$).

Based on Equations (4)-(8), the parameters $L_b$, $L_t$, $L_{tc}$, $M_b$, $M_t$ are calculated by data in Table II obtained from the anthropometry research [36]. This model is customizable because each individual's weight and height can be normalized by $M_W$ and $L_H$ respectively. $M_{sb}$ is the mass of the subject and the $L_{sb}$ is the height of the subject. $M_W$ is the total mass of the human model and $L_H$ is the total height of the human model from the anthropometry study.

$$M_b = (M_{sb}/M_W) \cdot \sum_{i=1}^{8} M_i \qquad (4)$$

$$M_t = (M_{sb}/M_W) \cdot M_9 \qquad (5)$$

$$L_b = (L_{sb}/L_H) \cdot \{[\sum_{i=1}^{8}(M_i \cdot L_i)/\sum_{i=1}^{8}(M_i)] - L_{12}\} \qquad (6)$$

$$L_t = (L_{sb}/L_H) \cdot (L_{12} - L_{13}) \qquad (7)$$

$$L_{tc} = (L_{sb}/L_H) \cdot (L_9 - L_{13}) \qquad (8)$$

TABLE II. THE HUMAN SEGMENT PARAMETERS

| # | Segment | $M_i$: Mass (Kg) Total Weight $M_W$: 81.4 Kg | $L_i$: Length between Center of Mass to Ground (m) Total Height $L_H$: 1.784 m |
|---|---|---|---|
| 1 | Head | $M_1$: 4.2 Kg | $L_1$: 1.679 m |
| 2 | Neck | $M_2$: 1.1 Kg | $L_2$: 1.545 m |
| 3 | Thorax | $M_3$: 24.9 Kg | $L_3$: 1.308 m |
| 4 | Abdomen | $M_4$: 2.4 Kg | $L_4$: 1.099 m |
| 5 | Pelvis | $M_5$: 11.8 Kg | $L_5$: 0.983 m |
| 6 | Arms | $M_6$: 4 Kg | $L_6$: 1.285 m |
| 7 | Forearms | $M_7$: 2.8 Kg | $L_7$: 1.027 m |
| 8 | Hands | $M_8$: 1 Kg | $L_8$: 0.792 m |
| 9 | Thighs | $M_9$: 19.6 Kg | $L_9$: 0.75 m |
| 10 | Calfs | $M_{10}$: 7.6 Kg | $L_{10}$: 0.33 m |
| 11 | Feet | $M_{11}$: 2 Kg | $L_{11}$: 0.028 m |
| 12 | Hip Pivot to Ground | | $L_{12}$: 0.946 m |
| 13 | Knee Pivot to Ground | | $L_{13}$: 0.505 m |

## B. Posture Detection and Low-Level Torque Control

Our biomechanics model-based control strategy adaptively assists the wearer for both squat and stoop. [13] used a predefined and fixed torque reference and it only worked with a stoop or squat motion instead of the adaptive and generic reference torque in our method. The high-level controller runs at 1K Hz and the torque loop proportional-integral-derivative (PID) controller is implemented to track the reference assistive torque. The low-level controller is implemented by the velocity loop PID which runs at 20K Hz, and the current PID control runs at 200K Hz. The sampling rate of the IMUs is 400 Hz. The three x-axes Euler angles of IMUs are represented as trunk angle $\theta_b$, thigh angle $\theta_t$, and shank angle $\theta_s$ and they are calibrated to zero degrees at the beginning of the experiment when the subject was instructed to stand straight. The knee angle $\theta_k$ and hip angle $\theta_h$ are calculated by equation (9) - (10) and the positive directions of knee and hip represent an extension.

$$\theta_k = \theta_t - \theta_s \quad (9)$$
$$\theta_h = \theta_t - \theta_b \quad (10)$$

## C. Experimental Procedure and Squat Assistant Control

Our study was approved by the City University of New York Institutional Review Board, and all methods were carried out in accordance with the approved study protocol. Three healthy subjects without musculoskeletal injuries followed a metronome to perform each squatting cycle in 8 seconds shown in Fig. 7 and repeat 5 times for each experiment. The knee angle $\theta_k$, desired assistive torque $\tau_r$, actual assistive torque, raw EMG signal, and the average of root-mean-square (RMS) EMG signal was used to analyze the resistance during the squatting in the unpowered condition and zero torque control. The torque control tracking error was analyzed in the experiments with 10% 30%, and 50 % of biological torque assistance.

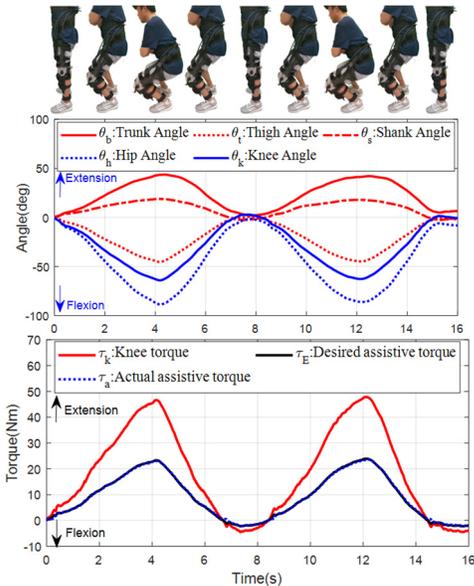

Fig. 7. The gesture detection and control strategy for the squat experiment with 50% of biological torque assistance. The top graph demonstrates the trunk, hip, thigh, knee, and shank angles with respect to time during squatting. The bottom graph depicts the calculated biological knee, desired assistive, and actual assistive torques.

## VI. EXPERIMENTAL RESULTS

To demonstrate the high-backdrivability characteristics, the control performance, and assistive performance, the results of the experiment of resistive torque in unpowered condition, the resistive torque in zero torque tracking control, the tracking performance in assistive control, and the evaluation of assistive control are described in this section.

### A. High Backdrivability in Unpowered Condition

To demonstrate high-backdrivability, the mechanical impedance (measured as the resistive torque) during the squatting in unpowered condition was studied. Thanks to quasi-direct drive actuation and low-friction cable-drive mechanism, it generated low impedance as the maximum resistive torque is 2.58 Nm which took place during the onsets of motor rotation and the changes of the direction. The backdrivability of our knee exoskeleton (2.58 Nm peak resistance) is superior to the knee exoskeleton [30] (8 Nm peak resistance). The average of the absolute resistance was 0.92 Nm, as shown in Fig. 8. Subjects reported extremely low resistance while wearing the device.

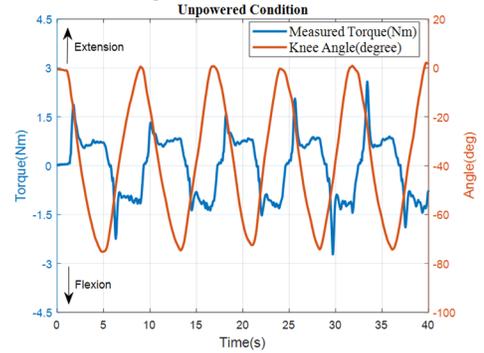

Fig. 8. The result of the mechanical impedance characterization during squatting. The maximum torque of the mechanical resistance is 2.58 Nm and the average torque is 0.92 Nm. It reveals that our robot is highly-backdrivable with low mechanical impedance [30].

### B. Zero Torque Tracking Control

The same subject performed squatting to further investigate the characteristics of mechanical impedance during the zero torque tracking control. Its reference torque was set to zero regardless of human motion. The zero torque control was implemented to eliminate the mechanical resistance, such as friction of the cables and gears.

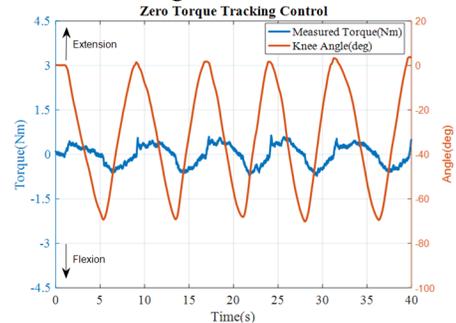

Fig. 9. The result of the zero torque tracking control. The reference torque is set to zero. It demonstrates that the maximum resistive torque is approximately 0.64 Nm and the average of the absolute measured torque is 0.34 Nm. It reveals that the zero torque tracking control further reduces the mechanical resistance on top of the intrinsic low impedance due to quasi-direct drive actuation.

In Fig. 9, the blue line indicates the measured torque and the orange line indicates the knee angle. The maximum measured torque (mechanical impedance) was approximately 0.64 Nm and the average of absolute measured torque was 0.34 Nm. Compared to the unpowered condition, the maximum torque in zero torque control was further reduced by 4 times and the average of the absolute measured torque in zero torque control was reduced 2.7 times. Therefore, it guarantees that the exoskeleton does not increase human energy consumption due to the mechanical resistance using the zero torque tracking control.

*C. Torque Tracking for Squatting Assistance*

Tests for 10%, 30%, and 50% of biological knee joint torque assistance were performed to investigate the tracking performance. The knee torque $\tau_k$ was calculated by equation (2) and the desired assistive torque $\tau_r$ was calculated by equation (3). The gain was set at 0.1, 0.3, and 0.5 respectively. The assistive control was used to augment human knee joints by applying specific torque according to the current trunk angle $\theta_b$ and thigh angle $\theta_t$ during a squat cycle. The angles of human segments were detected by the IMU sensors mounted on the trunk, thigh, and shank.

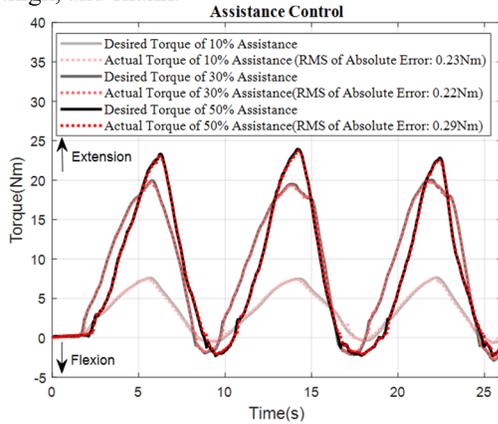

Fig. 10. The tracking performance of the 10%, 30%, 50% of knee torque assistance in three squatting cycles. The RMS of the absolute error between the desired and actual torque trajectory was 0.3 Nm, 0.22 Nm, and 0.29 Nm in 10%, 30%, and 50% knee assistance respectively. Root mean square (RMS) error of torque tracking was less than 0.29 Nm (1.21% error of 24 Nm peak torque).

The hysteresis and the backlash of the Bowden cable system cause major losses in force transmission and it affects the output torque, velocity and position control performance. [37] demonstrated that larger curvature angle increases the friction and, hence, a larger backlash width. Therefore, we minimized the deflection angle of the cable as small as possible to minimize the friction and applied a pretention force on the cable to minimize the backlash. The tracking performance is shown in Fig. 10. The RMS of the absolute error between the desired and actual torque trajectory was 0.23 Nm (2.8% of 7.6 Nm torque peak) in 10% knee assistance, 0.22 Nm (1.1% of 20 Nm peak torque) in 30% knee assistance, and 0.29 Nm (1.2% of 23.9 Nm peak torque) for 50% knee assistance. The torque tracking accuracy of (error is 1.2% of the desired peak torque) our biomechanics model-based control is superior to [38] (error is 2.1 Nm, 21% error of 10 Nm peak torque). It demonstrated that the torque controller can accurately deliver the desired torque profile to assist squatting.

Our tracking performance can be furthered improved by methods for hysteresis compensation in output position control by modeling the relationship between the input pulley angle and output pulley angle [39] or output torque control by modeling the relationship between the input pulley torque and output pulley torque [32, 40]. Nguyen et al. [32] proposed to compensate the nonlinearities and hysteresis effects from the cable conduit mechanism and control the output torque by modeling the relationship between the input pulley torque and the output pulley torque. In the future, a new adaptive control will be investigated to minimize the hysteresis and nonlinear properties of our exoskeleton system. It will also be an important and interesting research topic to characterize torque tracking performance between different hysteresis models [32, 37, 39, 40] using quasi-direct drive actuations.

*D. Injury Prevention Demonstration with EMG Sensors*

The effectiveness of muscle activity reduction using assistive control was evaluated in six robot loading scenarios. The knee extensors (rectus femoris, vastus lateralis, vastus medialis) and the knee flexor (biceps femoris and semitendinosus) are measured and there are three healthy male subjects (subject 1: 25 years, 170 cm, and 70 kg; subject 2: 32 years, 178 cm, and subject 3: 38 years, 175 cm, and 85 kg). We observed the amplitude of the raw EMG and the RMS value of absolute EMG which's RMS window is 0.1 second in Fig. 11. It shows that the muscle activities of vastus lateralis of subject 1 in six conditions (without wearing the exoskeleton, power-off exoskeleton, zero torque control, 10%, 30%, and 50% assistance). In passive condiction, EMG of power-off condition was slightly higher than the ones without-exoskeleton condition due to the passive mechanical resistance. In active condition, the EMG amplitude of the zero-torque was similar to the without-exoskeleton condition and the raw EMG and RMS EMG were reduced clearly in 10%, 30%, 50% assistance. It reveals that the assistive control reduced the muscle effort of knee extensor.

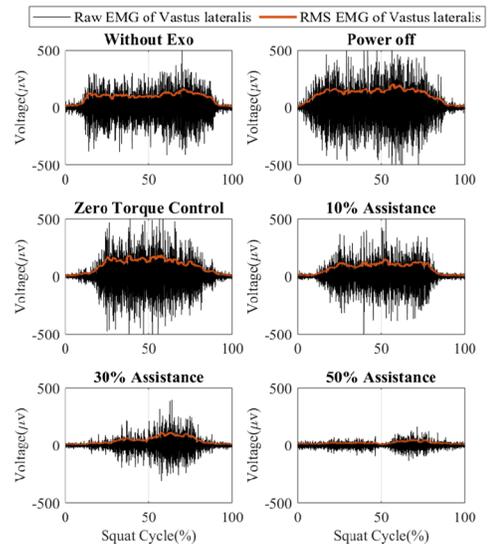

Fig. 11. Comparison of muscle activities of one subject between squatting in the conditions of without-exoskeleton, power-off, zero torque control, 10% assistance, 30% assistance, and 50% assistance. It shows that the amplitude of raw EMG with 50% assistance is smallest and it reveals that the assistive control reduced the muscle efforts in knee extensor vastus lateralis muscle.

Then, we averaged 15 squat cycles (5 squat cycles and 3 subjects) and observed the average amplitude of RMS EMG in five muscles (three knee extensors and two knee flexors) and six conditions to understand the whole assistive effect in three subjects. As shown in Fig. 12, it depicted that the more torque delivered to the subject, the more muscle activities of knee extensors (rectus femoris, vastus lateralis, vastus medialis) were reduced. The EMG of knee extensors had similar amplitudes in the conditions of without exoskeleton, power-off, zero torque control. But EMG in power-off condition had the highest amplitude. Compared to without-exoskeleton condition, peak EMG of the knee extensors (rectus femoris, vastus lateralis, and vastus medialis) in 50 % assistance were reduced by 87.5% (from 400 μV to 50 μV), 80% (from 500 μV to 100 μV) and 70% (from 500 μV to 150 μV) separately. However, we also observed that the muscle activities of knee flexor (biceps femoris and semitendinosus) slightly increased. This is possibly due to the lack of training of the exoskeleton device of those novice users. We will study if training and adaptation of the exoskeleton device may alleviate this minute side effects.

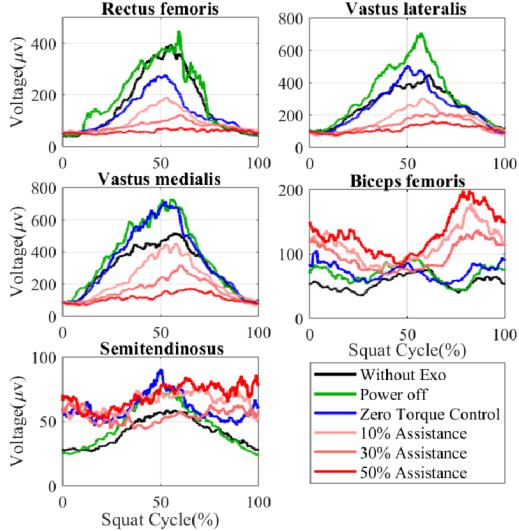

Fig. 12. Muscle activities of rectus femoris, vastus lateralis, (the major knee extensor muscle) under different assistive control levels. It shows the average of EMG in 15 squat cycles (three healthy subjects with 5 cycles each). Six conditions (without exoskeleton, power off, zero torque control, 10%, 30%, and 50% assistance) were compared. The result shows that the exoskeleton can obviously reduce the strength of the extensor with a slight increase in the strength of the flexor.

In Table II, compared to the without-exoskeleton condition, the average RMS EMG of knee extensor was reduced from 80 μV to 30 μV and the average RMS EMG of knee flexors are slightly increasing from 16 μV to 34 μV. The peak EMG of knee extensors in the power-off condition was 98 μV and it was slightly higher than 80 μV in the without-exoskeleton condition. It demonstrated that our exoskeleton still has mechanical resistance, but the mechanical resistance is fairly small and can be eliminated under the zero torque control. In summary, these results indicate that 1) the proposed exoskeleton is highly-backdrivable with minute mechanical resistance; 2) the assistance percentage should be large enough to reduce the muscle efforts and the squat with 10%, 30%, and 50% assistance can reduce the muscle effort effectively. We demonstrated that our proposed exoskeleton can reduce the knee extensors but it is still not clear that if the work is transferred to adjacent muscle groups (e.g. hip extensors, hip flexors, ankle extensors, and ankle flexors) under the exoskeleton assistance, In the future, the metabolics measurement will be used to enhance the analysis of the efficacy due to the complex mechanism of muscle group compensation.

TABLE III. THE AVERAGE OF RMS EMG IN KNEE MUSCLES

| | Muscles | Sub. | WO | Off | 0% | 10% | 30% | 50% |
|---|---|---|---|---|---|---|---|---|
| Knee Extensors | Rectus Femoris | S1 | 78 | 87 | 35 | 25 | 28 | 11 |
| | | S2 | 31 | 36 | 45 | 26 | 15 | 10 |
| | | S3 | 56 | 65 | 49 | 46 | 35 | 39 |
| | | Avg. | 55 | 62 | 43 | 32 | 26 | 20 |
| | Vastus Lateralis | S1 | 123 | 149 | 108 | 50 | 47 | 26 |
| | | S2 | 44 | 51 | 61 | 39 | 20 | 11 |
| | | S3 | 89 | 124 | 81 | 79 | 72 | 78 |
| | | Avg. | 85 | 108 | 83 | 56 | 46 | 38 |
| | Veastus Medialis | S1 | 87 | 124 | 111 | 50 | 46 | 14 |
| | | S2 | 98 | 117 | 135 | 78 | 38 | 12 |
| | | S3 | 112 | 129 | 101 | 98 | 88 | 89 |
| | | Avg. | 99 | 124 | 116 | 75 | 54 | 38 |
| | Knee Extensors | | 80 | 98 | 80 | 55 | 42 | 30 |
| Knee Flexors | Biceps Femoris | S1 | 23 | 25 | 23 | 63 | 43 | 57 |
| | | S2 | 13 | 20 | 18 | 13 | 16 | 20 |
| | | S3 | 20 | 25 | 32 | 37 | 37 | 55 |
| | | Avg. | 19 | 23 | 24 | 38 | 32 | 44 |
| | Semitendinosus | S1 | 13 | 14 | 11 | 20 | 15 | 18 |
| | | S2 | 14 | 19 | 19 | 14 | 14 | 15 |
| | | S3 | 15 | 14 | 30 | 31 | 24 | 36 |
| | | Avg. | 14 | 16 | 20 | 22 | 18 | 23 |
| | Knee Flexors | | 16 | 19 | 22 | 30 | 25 | 34 |

Unit (μV); WO (Without Exo); Off (Power off); 0% (Zero Torque); 10% (10% Assistance); 30% (30% Assistance); 50% (50% Assistance)

## VII. DISCUSSION AND CONCLUSION

This paper presents our endeavor to develop high-performance exoskeletons that minimize mass and stiffness, reduce the restriction of human movement, and enhance symbiotic control between human and robots. Weight minimization is achieved with high torque density motor and bidirectional cable drive using a single motor. The novel soft exoskeleton design mitigates high-pressure concentration (by maximizing moment arm of a soft robot) and reduces stiffness (quasi-direct drive actuation ensures high backdrivability). Compared with benchmarked exoskeletons, our design demonstrates high backdrivability (2.58 Nm peak resistance as compared to 8 Nm of a knee exoskeleton [30]) and high torque tracking accuracy (0.29 Nm, 1.2% of 23.9 Nm peak torque as compared to 2.1 Nm, 21% error of 10 Nm peak torque of a hip exoskeleton [38]). As proof of concept, the tethered exoskeleton demonstrates the design principles and effectiveness of control strategies. All design principles are transferable to portable version. Moreover, the offboard actuator is also lightweight. We are currently working on a portable exoskeleton design using the quasi-direct-drive actuation paradigm and high-torque-density motor. Human-exoskeleton interaction will be analyzed and discussed to optimize the wearable structures. Optimal control strategies will be investigated and the effectiveness of a portable version in the field will be studied using wearable motion and physiology sensors for injury prevention and human augmentation.